# Uncertainty Wrapper in the medical domain: Establishing transparent uncertainty quantification for opaque machine learning models in practice

Lisa Jöckel, Michael Kläs
*Data Science*
Fraunhofer IESE
Kaiserslautern, Germany
{lisa.joeckel, michael.klaes}@iese.fraunhofer.de

Georg Popp, Nadja Hilger, Stephan Fricke
*Department of Cell and Gene Therapy Development*
Fraunhofer IZI
Leipzig, Germany
{georg.popp, nadja.hilger, stephan.fricke}@izi.fraunhofer.de

*Abstract*— When systems use data-based models that are based on machine learning (ML), errors in their results cannot be ruled out. This is particularly critical if it remains unclear to the user how these models arrived at their decisions and if errors can have safety-relevant consequences, as is often the case in the medical field. In such cases, the use of dependable methods to quantify the uncertainty remaining in a result allows the user to make an informed decision about further usage and draw possible conclusions based on a given result. This paper demonstrates the applicability and practical utility of the Uncertainty Wrapper using flow cytometry as an application from the medical field that can benefit from the use of ML models in conjunction with dependable and transparent uncertainty quantification.

## I. INTRODUCTION

There are promising applications for artificial intelligence, especially machine learning, in the medical field. They can be used to build data-driven models (DDMs), which become part of the overall software system to provide functions that cannot be realized up to now by traditional software. However, as their behavior is derived from a data sample, correct DDM outcomes cannot be guaranteed. This becomes critical if errors can have safety-critical consequences. Furthermore, most DDMs today are opaque or even black-box models, i.e., they do not provide insight into how they determined the outcome.

Hence, model-agnostic uncertainty quantification approaches that provide situation-aware uncertainty quantification and are based on a solid statistical foundation are beneficial to complement the outcome of a DDM. The uncertainty wrapper (UW) framework [1] [2] was developed to provide such uncertainty quantification using an approach that is interpretable and can be checked by domain experts.

The application of UWs seems thus also promising in the medical context; however, applications published to date focus on the automotive domain. This raises the question of whether meaningful UWs can be built on data and models as applied in the medical domain and whether they can provide useful insights for related use cases.

In this paper, we try to tackle these questions in a case study on a medical application in which we illustrate the construction and evaluation of UWs on a black-box DDM for flow cytometry, which automatically classifies the cell populations of a blood sample based on marker intensities.

Contributions: We show (i) how uncertainty quantification can be implemented with UWs, including general steps to build UWs, deriving application-specific quality factors that are used by the UW, and showing their effect on uncertainty quantification performance. Moreover, we illustrate (ii) how the application of DDM in flow cytometry can benefit from uncertainty quantification at two levels: (a) getting upper- and lower-level boundaries on the relative amount of a specific cell type in a given blood sample, which indicates the situation-specific amount of uncertainty in the DDM results, and (b) including uncertainty information on an individual event base in gating plots to provide hints on the cause of observed uncertainty, allowing better informed decision-making regarding severity and potential countermeasures to be taken to reduce the observed uncertainty.

This paper is structured as follows: Section 2 provides background on the application of flow cytometry and DDMs applied in this context. Section 3 summarizes existing work on uncertainty quantification for DDMs. Section 4 presents our case study on the application of UWs for flow cytometry. Section 5 provides and discusses the study results, and Section 6 concludes the paper.

## II. BACKGROUND

This section gives a brief introduction to the application and to existing work on DDMs for flow cytometry.

### A. Application: Flow Cytometry

Flow cytometry is one of the major techniques for cell measurement and characterization. It is used in multiple fields of biomedicine, especially in diagnostics, for monitoring issues, or in research projects [3]. A major application of flow cytometry is in particular the characterization of Advanced Therapy Medicinal Products (ATMP), e.g. CAR T cells, whose development is progressing and represents a milestone in medicine [4].

To obtain a flow cytometric measurement, mainly blood or other suitable material (like bone marrow) are first collected from the patient. These *samples* are prepared and stained with multiple fluorescence-conjugated antibodies according to the panel specifications (see also Fig 1) [5]. The cells are analyzed one after the other through a thin measuring chamber (flow cell). This is achieved by hydrodynamic focusing. The cells are then illuminated with lasers, producing scattered light and fluorescence signals that are characteristic of each cell type. The signals are detected and stored as event-specific *marker intensities*.

The data generated is typically evaluated manually (referred to as *gating*) by visualizing marker intensities in hierarchically structured 2D plots and selecting cell populations based on known cell characteristics [6]. In the panel available here, cell populations were gated by placing quadrants into the 2D plots (see Fig. 1, step 4). Due to the dependency of the gating process on the operator, automated analysis techniques were introduced to overcome process and reproducibility bottlenecks [7].

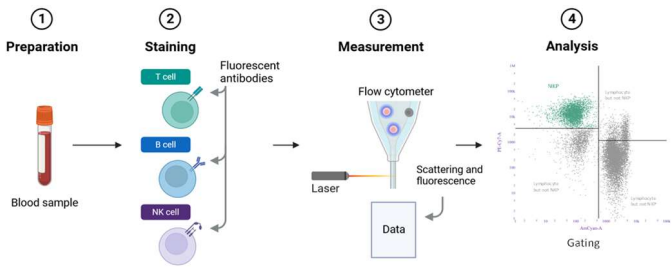

Fig. 1 Simplified process to obtain flow cytometric measurements, from preparation to analysis via manual gating.

### B. Data-driven Models for Flow Cytometry

Today, a variety of automated analysis techniques are available that have undergone extensive reviews [8] [9]. Due to the availability of large, labeled data sets from manual gating, supervised ML approaches have been used successfully [10] and are well suited to overcome translational hurdles in biomedicine, as they can be benchmarked against manual gating, which is still considered the gold standard. To contribute to the translational step, issues regarding unreliably classified cell events and their influence on aggregating the cell populations still need to be addressed. This is particularly important as operators need additional information about whether the model predictions are correct.

## III. RELATED WORK

The section provides a definition of uncertainty in the context of DDMs, introduces potential sources of DDM-related uncertainty, and provides an overview of model-intrinsic as well as externalized uncertainty quantification approaches.

### A. Uncertainty in Data-Driven Models

In the context of DDMs applied for classification tasks, a working definition for *uncertainty* is the likelihood that some information provided by the DDM is wrong.

The onion shell model distinguishes *sources of uncertainty* due to limitations in model fit, limitations in input quality, and scope compliance [11]. Uncertainty from model fit is related to the general capability of the DDM to provide reliable outcomes and is measured by DDM testing using adequate performance metrics. Input quality-related uncertainty can be estimated by modeling how the DDM performance changes depending on the input quality. Scope compliance uncertainty needs to be monitored as a DDM is always built and evaluated for a specific context of operation (i.e., intended use) and applications outside its scope can result in unforeseen failures.

### B. Model-Intrinsic Uncertainty Quantification

Sometimes, uncertainty quantification capabilities are realized as part of the DDM itself. In such cases, the DDM returns, for example, not only the outcome of interest, but also a probability for the outcome being correct, which we refer to as model-intrinsic uncertainty quantification.

A naïve approach to getting uncertainty quantification is to take the *overall error rate* on a test data set as an estimate for all outcomes. This is usually done through statistical testing and considers only model fit uncertainty.

Some classes of DDMs (e.g., support vector machines or logistic regression) provide *preference values* together with their outcome, which are commonly interpreted as certainties. These preference values, however, are mostly no real probabilities and were derived from training data, which tends to overconfident estimates. The preference values can be *calibrated* in a post-processing step to address these issues [12]; however, besides the DDM input, other information sources might be relevant for uncertainty quantification but are not considered yet. For deep neural networks, more advanced approaches are usually proposed such as deep ensembles or Bayesian neural networks [13].

### C. Externalized Uncertainty Quantification

Two externalized uncertainty quantification approaches that can provide uncertainty quantifications with statistical guarantees are Conformal Prediction [14] and Uncertainty Wrapper (UW) [1]. Conformal Prediction has its benefits when providing prediction regions instead of a single-point prediction, which contain the intended outcome with a given probability [15]. As we consider a binary classification task with uncertainty estimates for single-point predictions in this paper, we focus on UWs for uncertainty quantification.

UWs consider all three sources of uncertainty as described in the onion shell model. As shown in Fig. 2 [16], UWs contain a quality and a scope model, which model factors related to input quality and scope compliance, respectively, and might use additional information to the DDM input. For example, for a vision task outside using a camera, rain or darkness as input quality-related factors might help to differentiate the degree of uncertainty in different situations. Similarly, the location can be used to determine whether the DDM is still operating within its intended application scope. Therefore, additional information about the weather, time of day, or GPS coordinates can be used for uncertainty quantification.

The scope compliance-related factors are used to estimate the probability that the DDM is used outside its intended application scope. Based on the input quality-related factors, the application scope of the DDM is decomposed into areas of similar uncertainties using a decision-tree-based approach, referred to as quality impact model. Each leaf is associated with the uncertainty for an area of the application scope. To prevent underestimating the uncertainties corresponding to each leaf, a requested statistical confidence level is considered.

In previous work, a framework for building UWs was proposed [2] and the UW uncertainty quantification performance was compared to state-of-the-art model-intrinsic [17] as well as externalized uncertainty quantification approaches [15], with promising results.

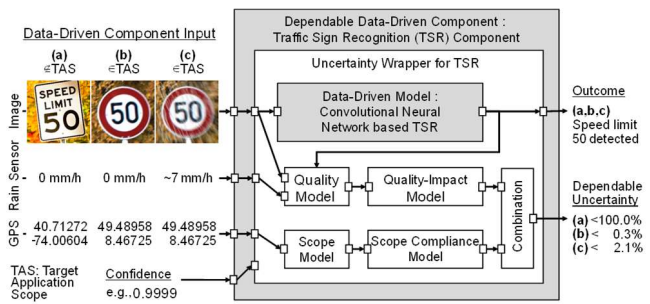

Fig. 2. Uncertainty Wrapper Framework.

In general, the UW approach is not specific to any application domain. It is possible to build UWs for various tasks by selecting and designing suitable factors related to input quality and scope compliance. However, previously published applications were in the automotive domain, mainly using image data for classification and detection tasks (e.g., [2] [17] [18] [19]). Here, we consider an application in the medical domain with numeric data and DDMs with a significantly lower error rate (i.e., model fit uncertainty).

## IV. Application for Flow Cytometry

In this section, we describe a case study on applying the UW framework in the context of flow cytometry. First, we will introduce the context and background of the study including the data sets and DDMs we considered. Next, we will explain the construction of the UWs step-by-step, giving details on the application-specific quality models and quality impact models. Finally, we will describe the application on *two use cases*, one at (a) the *cell population level* and one at (b) the *event level*.

### A. Study Context

Flow cytometry data was available for 100 healthy donors. The use of the data was approved by the local ethics committee and the respective informed consent of the subjects was available. The flow cytometric measurement was performed using an accredited clinical assay for immunophenotyping. The generated data was labeled using manual gating. Additionally, fluorescence values within the data were compensated and manual gating results were transformed into event-based class information.

For each donor, one individual sample of ~65k events was collected. The samples were randomly divided into training, calibration, and test data sets. The training data consisted of 2,170,486 events from 33 samples, the calibration data of 2,294,156 events from 33 samples, and the test data of 2,423,134 events from 34 samples.

We considered four cell types that should be correctly identified: lymphocytes ($L$) and three major lymphocyte subtypes: B cells ($BP$), T cells ($TP$), and natural killer cells ($NKP$). For each cell type, a small neural-network-based DDM was built that performed a binary classification task for each event $e_i$ based on its marker intensities $m_{i,1...n}$. For example, $\text{DDM}_L: f(m_{i,1...n}) = \hat{1}_L(e_i)$.

### B. Create Quality and Scope Model

Each of the DDMs was considered separately for uncertainty quantification, i.e., each UW encapsulated only one of the DDMs. The reason is that each of the DDMs had different performance and different factors influencing uncertainty.

We considered different sets of quality factors as input for the quality model of the UWs. The quality factors for a lymphocyte UW (i.e., a UW for the lymphocyte DDM) were built on all training events. The UWs for the subtypes only considered lymphocyte training events as a basis (which is the way a manual gating would also commonly be performed).

We considered five different groups of input quality-related factors for this application. A simple type of quality factors only used the ***marker*** intensities considered during gating for each cell type. The naïve assumption is that rather low or high intensities indicate a lower degree on uncertainty than intermediate values.

Three advanced quality factors were tailored to the aspect of the application that multiple events together form the decision basis for one blood sample. They were initialized during UW development but fit themselves to a new sample at runtime. Contrary to previously developed quality factors that consider only the data point itself, this new concept also incorporates the context of a data point, i.e., the blood sample as context of an event.

For ***density*** quality factors, a kernel density estimation function is fitted on events of the current sample for the pairwise marker intensities used for gating. In the gating plots, there are regions with many events clustered together and regions with no events. The idea is to exploit this for uncertainty quantification as the assumption is that different degrees of density are associated with different uncertainties. Another quality factor is based on the ***homogeneity*** of these clustered regions. Here, scikit-learn's DBSCAN algorithm [20] detects clusters of events for the sample and determines how similar an event is to the cluster it belongs to, i.e., the ratio of events with the same prediction as the event itself. This exploits the assumption that the more similar an event is to its cluster, the higher the probability that the DDM prediction was correct. For density- and homogeneity-based factors, a logarithmic transformation is performed on the marker intensities as the gating plots also use logarithmic scale. The third quality factor uses ***percentile*** ranks of the markers used during gating. An example of outcomes of these quality factors is depicted in Fig. 3.

The fifth kind of quality factor uses the DDM ***outcome*** directly, as the prediction performance of the DDM might vary between positive and negative events or other quality factors have a different interpretation depending on whether the DDM predicts a specific cell type or not.

We consider UW variants with different quality factor combinations. The outcome quality factor can be used optionally in addition to any of these combinations. The **basic** variant only uses the marker quality factors. The **density** variant uses density quality factors in addition to marker quality factors. Analogously, the **homogeneity** and **percentile** variants use respective quality factors in addition to marker quality factors. A further variant **combines** density, homogeneity, and marker quality factors as these are the most promising for the application. Please note that lymphocyte UWs do not use homogeneity quality factors, as the benefit of clustering is more promising for the subtypes.

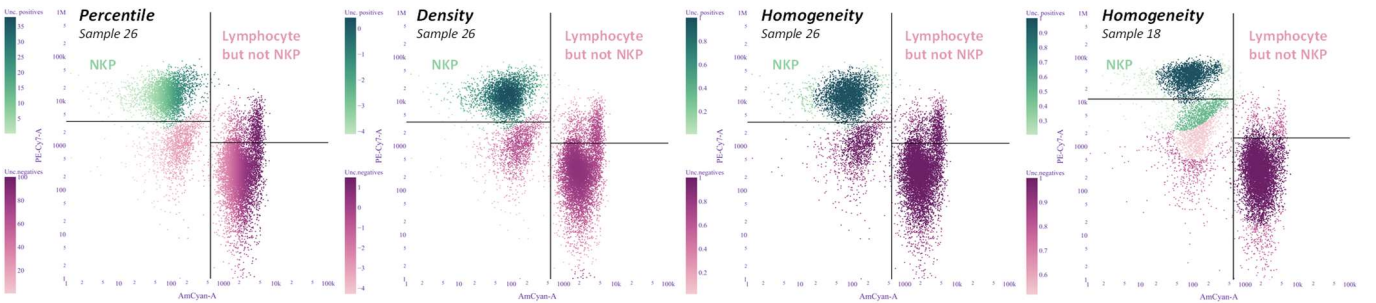

Fig. 3. Illustration of different kinds of quality factor outcomes associated with NKP cell events in gating plots for some examples. The UW uses these values as input to the decision tree-based quality impact model to determine the uncertainies per event.

The scope model was not explicitly implemented for this application as all samples were from the intended application scope. Scope factors might be derived from the common range of marker intensities, additional patient information like age and gender, or out-of-distribution-based scope factors for the marker intensities.

*C. Build and Calibrate Quality Impact Model*

We considered two different types of quality impact model for UWs, which include performance differences of the DDM depending on the event being predicted as the target cell type or not. The first type uses the **default quality impact model** but **includes the outcome quality factors** in the quality model. In the case of a lymphocyte UW, all training events are provided to the quality model (in the other cases, only the events predicted as lymphocytes), and its results are used as input to build the decision tree of the quality impact model. The second variant introduces a **category-based quality impact model** that adjusts the default quality impact model for this application. Here, events predicted as the target cell type are separated from those not predicted as the target cell type. Both sets of events are used to build one decision tree each. Fig. 4 shows an example excerpt of such a decision tree. Depending on the DDM prediction, the uncertainty determined from the corresponding decision tree is used as uncertainty estimate. For the category-based quality impact model, the **outcome quality factor is not included**. To establish a **baseline**, a UW with basic quality factors and the default quality impact model, but without outcome quality factor, is created. After training, the quality impact models were calibrated on the calibration data considering a confidence level of 0.99. Leaves containing fewer than 200 data points for lymphocyte UWs, and 50 data points for the other cell types, after calibration were pruned.

*D. Uncertainty Wrapper Application*

We considered two use cases where uncertainty quantification can be advantageous in the application. The first was to provide lower and upper bounds for the cell population, i.e., the cell type ratio at the **aggregation level** considering the event uncertainties. Information about the numerical cell ratios and their activation state allows medical experts to diagnose whether any disorders are present in the cellular immune system. A lower and upper bound for the number of *lymphocyte events* of a sample would be $|L|_{min} = \sum(cert(L_p))$, and $|L|_{max} = |L_p| + \sum unc(\neg L_p)$, respectively, where $L$ is the set of lymphocyte events, $L_p$ the set of events predicted as lymphocytes, $unc(X)$ the uncertainties associated with the events in $X$, and $cert(X) = 1 - unc(X)$ the certainty. As the lymphocyte event ratio is determined relative to the set of all events $E$ for the blood sample, we get $\left[\frac{|L|_{min}}{|E|}, \frac{|L|_{max}}{|E|}\right]$. For BP, TP and NKP events, a lower bound is $|C|_{min} = \sum\left(cert(C_p) \cdot cert(L_p)\right)$ with $C$ as the set of respective cell type events (i.e., BP, TP, or NKP), and $C_p$ the set of predicted cell type events. An upper bound is $|C|_{max} = |C_p| + \sum unc(\neg C_p) + \sum unc(\neg L_p)$. The ratio of BP, TP, and NKP is determined relative to the number of lymphocyte events: $\left[\frac{|C|_{min}}{|L|_{max}}, \frac{|C|_{max}}{|L|_{min}}\right]$.

The second use case was the integration of uncertainty estimates at the **event level** in the gating plots, thereby allowing experts to easily check the plausibility of the DDM for the current sample.

## V. RESULTS AND DISCUSSION

This section includes evaluation metrics, results, and discussion on our case study on UWs for the application of flow cytometry.

*A. Uncerainty Wrapper Performance Evaluation*

The UWs for the lymphocyte DDM were evaluated on all events in the test data. The UWs for the lymphocyte subtypes were evaluated on the subset of lymphocyte events in the test data. In all cases, uncertainty quantification performance is measured by the mean squared difference between the predicted probability associated with an outcome and the actual outcome, which is called *Brier score* [22]. A decomposition of the Brier score is $bs = var - res + unr$ [23]. The *variance var* relates to a high error rate of the DDM, i.e., model fit uncertainty. As $var$ bounds the *resolution res* and higher resolution values are better, we consider *unspecificity* as $var - res$ instead. The *unreliability unr* measures the calibration of the uncertainty estimates to the observed error rate of the DDM. In addition, we consider *overconfidence* as the part of unreliability where uncertainty quantification underestimates the observed error rate, which is more serious in this use case. For all considered evaluation metrics, smaller values indicate a better performance.

We compared uncertainty quantification performance at the event level of different UW variants using the Brier score and its components. As Table 1 shows, including the DDM outcome provide better estimates than the baseline. UWs with a category-based quality impact model sometimes outperformed their counterparts with the default quality impact model and the outcome quality factor. Depending on the cell type, different quality factors can contribute to a better uncertainty quantification performance. Lymphocyte UWs benefit from the density quality factor. For cell types like BP and TP, where the DDM already provides high prediction quality (i.e., low variance), quality factors used in addition to the marker quality factors did not yield any improvement. For NKP cells with a comparatively moderate DDM prediction quality, a combination of density- and homogeneity-based quality factors was best. The percentile quality factors do not appear to improve the uncertainty quantification.

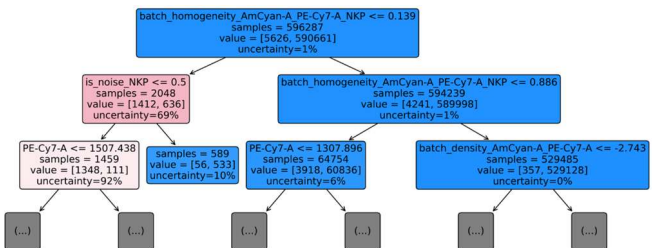

Fig. 4. Example of a category-based quality impact model for NKP events.

TABLE I. Performance results of UW variants.

| Variant | Brier sc. | Variance | Unspec. | Unreliab. | Overconf. |
|---|---|---|---|---|---|
| Baseline *Lym.* | .00311 | .00468 | .00290 | .00021 | .00002 |
| Basic *Lymphocyte* | .00286 | .00468 | .00270 | .00016 | .00003 |
|  | .00279* | .00468* | .00262* | .00018* | .00004* |
| Percentile *Lymphocyte* | .00292 | .00468 | .00271 | .00022 | .00003 |
|  | .00294* | .00468* | .00277* | .00018* | .00003* |
| Density *Lymphocyte* | .00259 | .00468 | .00250 | .00008 | .00003 |
|  | .00255* | .00468* | .00243* | .00012* | .00006* |
| Baseline *BP* | .00011 | .00182 | .00010 | .00001 | .00001 |
| Basic *BP* | .00010 | .00182 | .00009 | .00001 | .00001 |
|  | .00010* | .00182* | .00009* | .00001* | .00001* |
| Baseline *TP* | .00048 | .00097 | .00037 | .00001 | .00001 |
| Basic *TP* | .00039 | .00097 | .00030 | .00001 | .00001 |
|  | .00039* | .00097* | .00030* | .00001* | .00001* |
| Baseline *NKP* | .00511 | .00656 | .00424 | .00088 | .00014 |
| Basic *NKP* | .00486 | .00656 | .00396 | .00090 | .00025 |
|  | .00495* | .00656* | .00399* | .00096* | .00025* |
| Percentile *NKP* | .00588 | .00656 | .00410 | .00178 | .00047 |
|  | .00590* | .00656* | .00383* | .00207* | .00053* |
| Density *NKP* | .00462 | .00656 | .00394 | .00068 | .00009 |
|  | .00487* | .00656* | .00382* | .00105* | .00022* |
| Homogeneity *NKP* | .00506 | .00656 | .00370 | .00137 | .00054 |
|  | .00439* | .00656* | .00373* | .00065* | .00024* |
| Combined *NKP* | .00438 | .00656 | .00402 | .00036 | .00010 |
|  | .00429* | .00656* | .00374* | .00055* | .00024* |

\* Category-based quality impact model

### B. Use Case 1: Uncertainty at the Aggregation Level

The UWs extend the cell population predictions for a blood sample, which are based on the corresponding DDM outcomes, with lower and upper uncertainty boundaries on the chosen confidence level of 0.99. Fig. 5 depicts for each sample in the test dataset the true cell population (in increasing order), the predicted population, and the UW-based boundaries.

We can observe that for cell types with high prediction accuracy, we get rather small uncertainty ranges (e.g., BP). For samples with visible deviations between the predicted and actual population, there is a clear tendency to wider uncertainty ranges compared to other samples. This means the uncertainty quantifications provided by the UWs can help the user to identify suspicious DDM prediction results and check them at the event level (cf. use case 2). In the next subsection, we will illustrate this for NKP on sample 26, 28, and 18.

Moreover, in most cases the uncertainty ranges tend to be wider in the direction in which the true population deviates from the predicted one and commonly also include the true population. However, in some case, the uncertainty range is still too narrow (e.g., sample 18), which might be due to violations of the independence assumption for the data points we used for calibration since events from the same sample are inherently dependent due to their origin. In practice, it is however nearly impossible to obtain independent event data without collecting and analyzing an unrealistic high number of blood samples.

### C. Use Case 2: Uncertainty at the Event Level

Uncertainty results on the event level are visualized using the plots that would also be used in manual gating (Fig. 6). Events predicted to belong to the cell population of interest (in our example selection NKP) are colored in green, and other events (in this case predicted as lymphocytes but not NKP) in purple. Darker colors indicate higher uncertainty. Black lines are the actual gates from manual gating. They are usually not available for new samples but used in this case to illustrate the ground truth. All cells in the upper left quadrant have been identified as NKP cells by the manual gating activity.

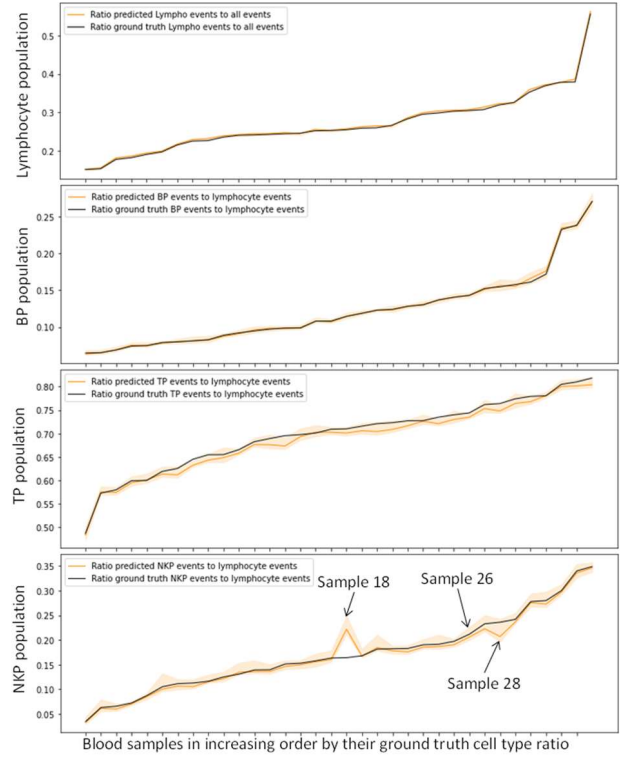

Fig. 5. Cell population boundaries based on aggregated event uncertainties.

Next, we use sample 26, 28, and 18 to illustrate how the uncertainties on the event level can help users to make more informed decisions. For sample 26, we see in Fig. 6 that the DDM predicts the cluster of NKP as we would expect, which however differs in this case from the ground truth provided by the less accurate manual gating. The uncertainty estimates highlight the event in this area as highly uncertain since the manual gating would commonly identify them as NKP although they are not. Checking the source of the wide uncertainty range on the population level (cf. use case 1) on event level would in such cases lead to accept the prediction result of the DDM.

Checking the events with high uncertainty for sample 28 highlights that there is an intersection of cell populations in the sample that could not be well separated by the selected combination of markers. To reduce uncertainty, a new blood sample must be taken, or a different set of markers selected that improve the separation between NKP and non-NKP. The uncertainty is related to the event data which are used as input and not the performance of the DDM. Thus, it could also not be significantly reduced by a follow-up manual gating.

For sample 18, there is actually a problem with the DDM outcomes. The UW did not catch all events with high uncertainty, but at least indicates the cluster where the DDM is incorrect. The previously mentioned dependencies in the calibration data and the chosen confidence level might have contributed to the fact that the UW did not catch the full range of the problem for this sample. In this case a follow-up manual gating can reduce the uncertainty for the sample.

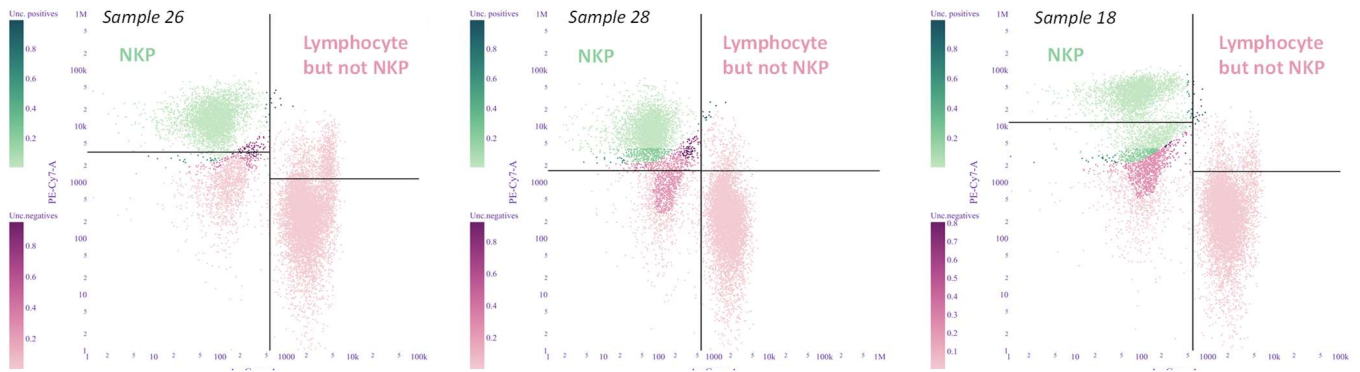

Fig. 6. Illustration of event level uncertainty in gating plots for selected test data samples using the uncertainty wrapper in the combined variant.

## VI. Conclusions

The bottlenecks of manual gating in flow cytometry analysis are well described and quantified [9]. Despite the availability of automated gating techniques and numerous reports of their benefit when used, the uptake of these methods in practice remains low [21]. In the foreseeable future, users will still need to verify the correctness of the results in flow cytometry due to the high requirements of healthcare services and products. Therefore, any DDM designed to support fast and objective decision-making (e.g., where to place a gate) needs to provide information about the quality of its prediction. We demonstrated the applicability of the Uncertainty Wrapper framework for the application of flow cytometry, hence showing promise for further applications in the medical field. To this end, we introduced new types of quality factors and a new type of quality impact model to derive the uncertainty quantifications. Here, we recommend considering the DDM outcome either as a quality factor or by splitting the quality impact model into two separate models using the introduced category-based quality impact model. For DDMs with very high prediction quality, basic quality factors based on the marker intensities are sufficient. Otherwise, including application-specific quality factors brings further improvement.

Furthermore, we demonstrated how DDMs for flow cytometry can benefit from uncertainty quantification: (1) by deriving lower and upper boundaries of cell populations based on aggregated event uncertainties, and (2) by visualizing event uncertainties in gating plots. The second case can contribute to finding issues with DDM-based gating (due to model limitations) or with the ground truth from manual gating (due to difficult positioning of the gates). Future directions include uncertainty quantifications for further medical applications and incorporating the findings into a structured quality assurance framework for DDMs.

### Acknowledgment

Parts of this work have been funded by the project "AIControl" as part of the funding program "KMU akut" of the Fraunhofer-Gesellschaft, and the project "RespiVir" as part of the Fraunhofer Center for Digital Diagnostics ZDD®. We would like to thank Prof. Dr. Ulrich Sack and Dr. Andreas Boldt from the Institute for Clinical Immunology, University Hospital Leipzig, for providing the flow cytometry data used in this study and their helpful discussions.